\documentclass[10pt,twocolumn,letterpaper]{article}

\usepackage{cvpr}
\usepackage{times}
\usepackage{epsfig}
\usepackage{graphicx}
\usepackage{amsmath}
\usepackage{amssymb}

\usepackage{subcaption}
\usepackage{multicol}

\usepackage{listings}
\usepackage{xcolor}
\definecolor{codegreen}{rgb}{0,0.6,0}
\definecolor{codegray}{rgb}{0.5,0.5,0.5}
\definecolor{codepurple}{rgb}{0.58,0,0.82}
\definecolor{backcolour}{rgb}{0.95,0.95,0.92}

\lstdefinestyle{mystyle}{
    backgroundcolor=\color{backcolour},   
    commentstyle=\color{codegreen},
    keywordstyle=\color{magenta},
    numberstyle=\tiny\color{codegray},
    stringstyle=\color{codepurple},
    basicstyle=\ttfamily\footnotesize,
    breakatwhitespace=false,         
    breaklines=true,                 
    captionpos=b,                    
    keepspaces=true,                 
    numbers=left,                    
    numbersep=5pt,                  
    showspaces=false,                
    showstringspaces=false,
    showtabs=false,                  
    tabsize=2
}

\allowdisplaybreaks

\usepackage[breaklinks=true,bookmarks=false]{hyperref}

\cvprfinalcopy 

\def\cvprPaperID{566} 

\begin{document}

\title{Likelihood Assignment for Out-of-Distribution Inputs in Deep Generative Models is Sensitive to Prior Distribution Choice}

\author{Ryo Kamoi, Kei Kobayashi\\
Keio University, Japan \\
{\tt\small ryo\_kamoi\_st@keio.jp, kei@math.keio.ac.jp}
}

\maketitle

\begin{abstract}
Recent work
has shown that deep generative models assign higher likelihood
to out-of-distribution inputs than to training data.
We show that a factor underlying this phenomenon
is a mismatch between the nature of
the prior distribution and that of the data distribution,
a problem found in widely used deep generative models
such as VAEs and Glow.
While a typical choice for a prior distribution is a standard Gaussian distribution,
properties of distributions of
real data sets may not be consistent with
a unimodal prior distribution.
This paper focuses on the relationship between
the choice of a prior distribution and the likelihoods assigned to
out-of-distribution inputs.
We propose the use of a mixture distribution as a prior 
to make likelihoods assigned by deep generative models 
sensitive to out-of-distribution inputs.
Furthermore, we explain the theoretical advantages of adopting a mixture distribution
as the prior,
and we present experimental results to support our claims.
Finally, we demonstrate that
a mixture prior lowers the out-of-distribution likelihood
with respect to two pairs of real image data sets:
Fashion-MNIST vs.\ MNIST and CIFAR10 vs.\ SVHN.
\end{abstract}

\section{Introduction}
The out-of-distribution detection is an important area of study
that has attracted considerable attention
\cite{Pimentel2014ADetection, Hendrycks2017a, Liang2018EnhancingNetworks, Shafaei2018ADetectors}
to improve the safety and reliability of machine learning systems.
Detection methods based on density estimation using a parametric model
have been studied for low dimensional data \cite{Pimentel2014ADetection},
and deep generative models seem to be a reasonable choice
when dealing with high-dimensional data.
However,
recent work \cite{Nalisnick2019, Hendrycks2019DeepExposure, Shafaei2018ADetectors, Nalisnick2019DetectingTypicality, Choi2019WAICDetection}
has shown that deep generative models such as VAEs \cite{Kingma_Auto_2013},
PixelCNN \cite{VanDenOord2016conditional},
and flow-based models \cite{Dinh2015NICE:Estimation, Kingma2018Glow:Convolutions}
cannot distinguish training data from out-of-distribution inputs
in terms of the likelihood.
For instance,
deep generative models trained on Fashion-MNIST assign
higher likelihoods to MNIST than to Fashion-MNIST,
and those trained on CIFAR-10 assign higher likelihood to SVHN 
than to CIFAR-10 \cite{Nalisnick2019}.
Methods for mitigating this problem
have been proposed from various perspectives
\cite{Hendrycks2019DeepExposure, Butepage2019ModelingModels, Choi2019WAICDetection, Nalisnick2019DetectingTypicality}.

We focus on the influence of the prior distribution of
deep generative models on the likelihood assigned to out-of-distribution data.
Although the typical choice is a standard normal distribution,
various studies have analyzed alternatives
\cite{Dilokthanakul_Deep_2016, Chen2017VariationalAutoencoder, Tomczak_VAE_2017, van_den_Oord_Neural_2017}.
However,
present work mainly focuses on the representative ability and
the likelihood assigned to in-distribution data
when evaluating prior distributions.
To the best of our knowledge,
no existing work has analyzed the effect
that the prior distribution has
on the likelihood assigned to out-of-distribution inputs.
Here, we consider data sets that can be naturally partitioned into clusters,
so the underlying distribution can be approximated by a multimodal distribution
with modes apart from each other.
This assumption is reasonable for many data sets found in the wild such as Fashion-MNIST,
which contains different types of images,
including T-shirts, shoes, and bags.
If a unimodal prior distribution is used to train generative models on such data sets,
the models are forced to learn the mapping between
unimodal and multimodal distributions.
We consider this inconsistency
an important factor underlying the assignment of high likelihood to out-of-distribution areas.

We use untrainable mixture prior distributions
and manually allocate similar data to each component
before training by using labels of data sets or k-means clustering.
Under these conditions, the models trained on Fashion-MNIST successfully assign
lower likelihoods to MNIST.
Our approach also lowers the likelihoods assigned to SVHN
by models trained on CIFAR-10.
We provide three explanations for our observations.
First, as mentioned above, a multimodal prior distribution can alleviate
the inconsistency between a prior and a data distribution,
which is a possible factor underlying the out-of-distribution problem.
Second, allocating similar data to each component
can reduce the possibility of accidentally assigning undesirable out-of-distribution points
to high likelihood areas.
Our second order analysis can theoretically justify
this intuition in a manner similar to the work of Nalisnick \etal \cite{Nalisnick2019}.
Third, out-of-distribution points are
forced out of high likelihood areas of the prior distribution
when a multimodal prior is used.
Somewhat surprisingly, the out-of-distribution
phenomenon still occurs when a model with a unimodal prior is
trained only on data that would be allocated to one component in the
multimodal case.
This is a novel observation that motivates
further investigation of designing the latent variable space to mitigate the out-of-distribution phenomenon.
\section{Related Work}

Our work is directly motivated by the recent observation
that deep generative models can assign higher likelihoods to
out-of-distribution inputs \cite{Nalisnick2019, Choi2019WAICDetection}.
The use of prior distributions
has been studied independently of this line of work.

\subsection{Out-of-Distribution Detection by Deep Generative Models}
Although model likelihood is often used to evaluate deep generative models,
Theis \etal \cite{Theis2016AModels} showed that
high likelihood is neither sufficient nor necessary for
models to generate high quality images.
Remarkably, Nalisnick \etal \cite{Nalisnick2019} has reported that 
deep generative models such as VAEs, flow-based models, and PixelCNN
can assign higher likelihoods to out-of-distribution inputs.
Similar phenomena have also been reported in parallel studies
\cite{Choi2019WAICDetection, Hendrycks2019DeepExposure}.

Solutions have been proposed from various perspectives.
Hendrycks \etal \cite{Hendrycks2019DeepExposure} 
proposed ``outlier exposure'',
a technique that uses carefully chosen outlier data sets during training to
lower the likelihood assigned to out-of-distribution inputs.
B\"utepage \etal \cite{Butepage2019ModelingModels} focused on VAEs
and reported that the methods for evaluating likelihood
and the assumption of a visual distribution on pixels
influence the likelihood assigned to out-of-distribution inputs.
Another line of study is to use alternative metrics.
Choi \etal \cite{Choi2019WAICDetection} proposed using
the Watanabe-Akaike Information Criterion (WAIC) as an alternative.
Nalisnick \etal \cite{Nalisnick2019DetectingTypicality}
hypothesized that out-of-distribution points
are not located in the model's ``typical set'',
and thus proposed the use of a hypothesis test
to check whether an input resides in the model's typical set.

\subsection{Prior distribution}
A typical choice for a prior distribution for deep generative models such as
VAEs and flow-based models is a standard Gaussian distribution.
However, various studies have proposed the use of different alternatives.
One line of study selects more expressive prior distributions,
such as multimodal distributions
\cite{Johnson_Composing_2016, Dilokthanakul_Deep_2016, Tomczak_VAE_2017, Nalisnick_Stick_2017, Izmailov2019Semi-SupervisedFlows},
stochastic processes \cite{Nalisnick_Stick_2017, Goyal2017a, Casale2018GaussianAutoencoders}, 
and autoregressive models
\cite{Chen2017VariationalAutoencoder, van_den_Oord_Neural_2017}.
Another option is to use discrete latent variables
\cite{Rolfe_Discrete_2017, van_den_Oord_Neural_2017}.
Previous work on the choice of the prior distribution for deep generative models
have focused on the representative ability, natural fit to data sets,
and the likelihood or reconstruction of in-distribution inputs.
To the best of our knowledge,
no previous study has focused on the relationships between
the prior distribution and the likelihood assigned to out-of-distribution data.

\section{Motivation \label{section:method}}
In this section, we discuss the theoretical motivations
for using a multimodal prior distribution;
topology mismatch and second order analysis.
On a related note,
we have observed that a multimodal prior distribution
can force out-of-distribution points out of
high likelihood areas.
We explain this effect in Section~\ref{subsection:twolabels}.

\subsection{Topology Mismatch \label{subsection:motivation}}

\begin{figure}[tb]
\center
\includegraphics[width=0.7\linewidth]{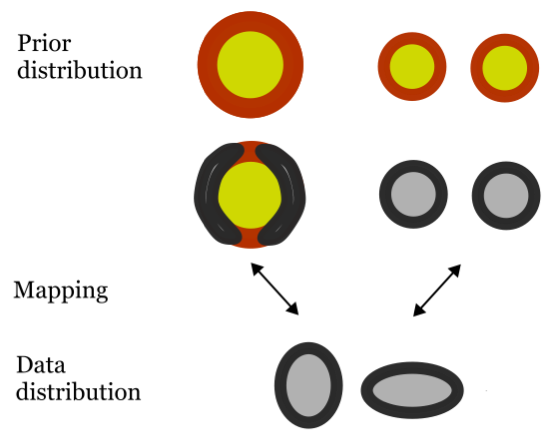}
\caption{Motivation for using a multimodal prior distribution from 
a topological point of view.
If the prior distribution is mapped to
a distribution with a different topology,
the mapped distribution will inevitably have undesirable high likelihood areas.
The black and red areas represent the typical sets of 
the prior and the data distribution, respectively.
The gray and yellow areas represent high likelihood areas of 
the prior and the data distribution, respectively.
While the distributions are shown in two-dimensions in this figure,
this inconsistency between high likelihood areas and
typical sets is a problem observed in high dimensional data.
\label{fig:topology}}
\end{figure}

\begin{figure*}[tb]
\begin{subfigure}[t]{0.33\textwidth}
    \center
    \includegraphics[height=3.1cm]{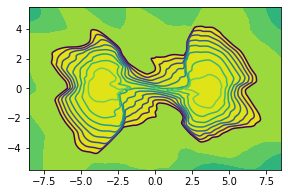}
    \caption{Standard Gaussian Prior \label{fig:simple_stdprior}}
\end{subfigure}
\begin{subfigure}[t]{0.33\textwidth}
    \center
    \includegraphics[height=3.1cm]{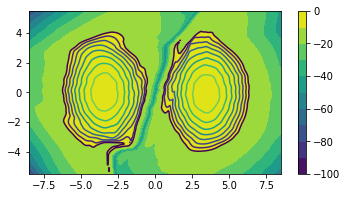}
    \caption{bimodal Gaussian Prior \label{fig:simple_gmm}}
\end{subfigure}
\begin{subfigure}[t]{0.33\textwidth}
    \center
    \includegraphics[height=3.1cm]{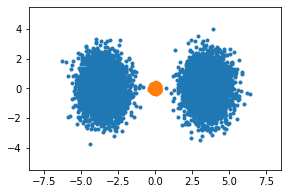}
    \caption{Data Points \label{fig:simpledata_visualize}}
\end{subfigure}
\caption{Visualization of the topology mismatch problem on a two-dimensional Gaussian mixture data.
(\subref{fig:simple_stdprior}, \subref{fig:simple_gmm})
Contours of the log-likelihoods assigned by
flow-based generative models using a standard Gaussian prior and
a bimodal Gaussian mixture prior.
The 10 contour lines in the images range from -10 to -1.
The model with a standard Gaussian prior assigns high likelihoods
outside the high probability areas of the true distribution.
(\subref{fig:simpledata_visualize})
Training data (blue) and out-of-distribution inputs (orange)
used in this experiment.}
\label{fig:simpledata}
\end{figure*}

We focus on data sets that have ``clusters'',
and adopt an assumption that the underlying distribution
can be approximated as a multimodal distribution
with components located far away from each other.
We analyze deep generative models by approximating them as
topology-preserving invertible mappings between
a prior distribution and a data distribution.
Nalisnick \etal \cite{Nalisnick2019DetectingTypicality} focused on
the ``typical set'' \cite{Cover2006ElementsTheory}
of deep generative models and the data distribution.
As suggested by Nalisnick \etal \cite{Nalisnick2019DetectingTypicality},
here, we assume that deep generative models learn
mappings from the typical set of the prior distribution to 
the typical set of the data distribution.
Figure~\ref{fig:topology} visualizes the intuition of
the mappings from a bimodal data distribution to
two different types of prior distributions under our assumptions.
If the bimodal data distribution is mapped to a unimodal prior distribution,
we cannot eliminate the possibility of the model mapping
out-of-distribution inputs to
the typical set or high likelihood areas of the prior distribution.
We will refer to this issue as the topology mismatch problem.
This simple analysis explains the out-of-distribution phenomenon
and the results of prior work \cite{Nalisnick2019DetectingTypicality},
implying that out-of-distribution inputs can even reside in the typical set.
By contrast,
if a prior distribution is topologically consistent with the data distribution,
there exists a mapping that decreases the possibility
of the out-of-distribution phenomenon.
Note that
we cannot say that the modification in a prior distribution can single-handedly solve
the problem as
the probability density of latent variables on
a prior distribution is not the only factor influencing
the likelihood of deep generative models
such as VAEs and Glow.
In addition,
it has been reported that deep generative models trained on
similar images can generate dissimilar images
\cite{Perera2019OCGAN:Representations},
and thus our analysis on topology mismatch cannot explain this result.
However, we later experimentally show that the choice of the prior distribution nonetheless
has a significant influence on the likelihoods assigned to out-of-distribution inputs
(Section~\ref{section:experiments}).

To justify our analysis,
we conduct experiments on some simple artificial data sets.
Figure~\ref{fig:simpledata} shows the likelihoods assigned by
flow-based deep generative models trained on points sampled from
a bimodal Gaussian mixture distribution.
We use a simple model architecture with four affine coupling layers
and reverse the features after each layer.
We compare a unimodal Gaussian prior and a bimodal Gaussian mixture prior.
Figure~\ref{fig:simpledata} shows
that the contours of the log-likelihood assigned by
the model using 
a standard Gaussian prior distribution 
have high likelihood areas outside the region data points reside.
Because the prior distribution is mapped to
a distribution with a different topology,
the mapped distribution will inevitably have undesirable high likelihood areas.
By contrast, the contours of the model using a Gaussian mixture prior
successfully separates the two modes,
and do not have high likelihood areas in out-of-distribution areas.
To show that a model with a standard Gaussian prior can assign
high likelihood to out-of-distribution inputs even in the low-dimensional case,
we compare the likelihoods assigned to out-of-distribution inputs
that are points sampled from a Gaussian distribution 
with mean zero and variance $0.01$.
As shown in Figure~\ref{fig:simpledata_visualize},
the out-of-distribution inputs have minimal overlap with 
the in-distribution data.
However, the mean of the log-likelihoods assigned by
the model using a standard Gaussian prior to in-distribution inputs is $-3.25$,
which is similar to
the log-likelihood assigned to out-of-distribution inputs ($-4.87$).
By contrast, the mean of log-likelihood assigned to out-of-distribution inputs
by the model using a multimodal prior is much lower ($-40.11$).
As the dimensionality of the data increases,
this phenomenon becomes more pronounced;
however, using a multimodal distribution as a prior can
significantly alleviated this problem.
Further details are presented in Appendix~\ref{appendix:artificial}.

\subsection{Second Order Analysis \label{subsection:secondorder}}
Nalisnick \etal \cite{Nalisnick2019} provided a second order analysis
with implications consistent with their experimental observations,
while they put some strong assumptions.
One implication of their analysis is
that deep generative models with unimodal prior distributions
assign higher likelihood
if out-of-distribution images have lower variance over image pixels.
However,
since they use a unimodal prior distribution,
their analysis do not apply here.
To explain why our proposition may help,
we perform a similar analysis on assumptions corresponding to our models.
Although we still adopt some strong assumptions and apply coarse approximations
in a similar manner as the original analysis,
our analysis provides an intuitive explanation for our experimental results.
The value we are interested in evaluating is 
\begin{equation} \label{eq:diff_of_likelihoods}
    \mathbb E_{q}[\log p(\mathbf x; \theta)] - \mathbb E_{p^*}[\log p(\mathbf x; \theta)]
\end{equation}
where $p$ is a given generative model,
$q$ is the adversarial distribution (out-of-distribution),
and $p^*$ is the training distribution.
If the value of is positive, the adversarial distribution is
assigned a higher likelihood by the generative model.
In the following analysis, we assume that $p$ reasonably approximates $p^*$.
Note that the analysis for unimodal prior models suggests that
$q$ can be assigned higher likelihood
even if $p$ perfectly approximates $p^*$.

Nalisnick \etal \cite{Nalisnick2019} approximate
the probability distribution function of
the given generative model $p$ as
$\log p(\mathbf x; \theta) \simeq \log p(\mathbf x_0; \theta)
+ \nabla_{x_0} \log p(\mathbf x_0; \theta)^T(\mathbf x-\mathbf x_0)
+ \frac{1}{2} \text{Tr} \{\nabla^2_{x_0} \log p(\mathbf x_0; \theta)(\mathbf x-\mathbf x_0)(\mathbf x-\mathbf x_0)^T \}$,
which is equivalent to assuming that
the generative model can be approximated with a Gaussian distribution.
In this work,
however, we focus on data sets with an underlying distribution
which can be approximated by a mixture distribution.
Therefore, we assume that $p$ can be approximated as
$\log p(\mathbf x; \theta) \simeq \log \frac{1}{K}\sum_K p_i(\mathbf x; \theta)$,
where $p_i$ corresponds to each component approximated by a Gaussian distribution.
We assume that each component of the generative model $p_i(\mathbf x; \theta)$
corresponds to a component of the prior distribution $p_i(\mathbf z; \psi)$.

Here, we adopt the assumption that the generative model is constant-volume Glow (CV-Glow)
as is done in \cite{Nalisnick2019}.
The derivation and detailed assumptions are given in
Appendix~\ref{appendix:secondorder}.
Finally, we derive the following formula:
\begin{align} \label{eq:final_formula}
    & \mathbb E_{q}[\log p(\mathbf x; \theta)] - \mathbb E_{p^*}[\log p(\mathbf x; \theta)] 
    \nonumber \\
    &\simeq -\frac{1}{2\sigma^2_{\psi}}
        \sum_{c=1}^C \left(\prod_{l=1}^L \sum_{j=1}^{C_l} u_{l, c, j} \right)^2
        \nonumber \\
    &\hspace{3em}
        \sum_{h, w} \sum_{i=1}^K (w_i \sigma^2_{\mathcal D_i, h, w, c}
        - w^*_i \sigma^2_{\mathcal D^*_i, h, w, c}) \nonumber \\
    &\le -\frac{1}{2\sigma^2_{\psi}}
        \sum_{c=1}^C \left(\prod_{l=1}^L \sum_{j=1}^{C_l} u_{l, c, j} \right)^2
        \nonumber \\
    &\hspace{3em}
        \sum_{h, w} (\sigma^2_{\mathcal D_{min}, h, w, c}
        - \sigma^2_{\mathcal D^*_{max}, h, w, c}),
\end{align}
where $\sigma^2_{\psi}$ is the variance of the prior distribution
(we assume that all the components have identical variance),
$\mathcal D^*_i$ and $\mathcal D_i$ correspond to 
the in-distribution and out-of-distribution data allocated to the $i$-th component
and $w^*_i, w_i$ are the ratio of data allocated to the $i$-th component
satisfying $\sum_K w^*_i = 1, \sum_K w_i = 1$.
$u_{l, c, j}$ is the weight of the $l$-th 1x1 convolution, which is fixed for any inputs.
Further, $h$ and $w$ index the input spatial dimensions,
$c$ indexes the input channel dimensions,
$l$ indexes the series of flows,
and $j$ indexes the column dimensions of the $C_l \times C_l$ kernel.
$\sigma^2_{\mathcal D_i, h, w, c}, \sigma^2_{\mathcal D^*_i, h, w, c}$ are diagonal elements of
$\Sigma_{\mathcal D_i} =
\mathbb E[(\mathbf x-\mathbf{\bar{x}}_i)(\mathbf x-\mathbf{\bar{x}}_i)^T | \mathcal D_i],
\Sigma_{\mathcal D^*_i} =
\mathbb E[(\mathbf x-\mathbf{\bar{x}}_i)(\mathbf x-\mathbf{\bar{x}}_i)^T | \mathcal D^*_i]$,
where $\mathbf{\bar{x}}_i$ is the elementwise mean of 
the images generated from the $i$-th component,
and the two matrices are assumed to be diagonal.
$\mathcal D_{min}$ and $\mathcal D^*_{max}$ are chosen so the final expression is maximized.
Expanding the formula by using CV-Glow does not seem to be a reasonable,
as we do not use it in our experiments.
However, Nalisnick \etal \cite{Nalisnick2019} reported that 
the out-of-distribution phenomenon occurs even on CV-Glow,
one of the simplest deep generative models,
as with many other more complex deep generative models such as general Glow and VAEs.
Therefore, it is worth considering CV-Glow to analyze the problem of
general deep generative models.

Roughly speaking,
we can say that if $\sigma_{\mathcal D_{min}}$ takes smaller values
than $\sigma_{\mathcal D^*_{max}}$, 
the likelihood assigned to out-of-distribution data
can be larger than that assigned to in-distribution data.
However, if this is the case, this indicates that 
one of the out-of-distribution modes has a mean
that is close to the mean of one of the modes of 
the generative model with small variance.
If out-of-distribution data satisfies this condition, 
such a mode can no longer be considered out-of-distribution, 
as inputs corresponding to it must be similar to 
the images corresponding to the mode of the generative model.
Note that a mode of the generative model has a small variance
and contains similar images under our assumptions.
By contrast,
the analysis by Nalisnick \etal \cite{Nalisnick2019}
assumed that the data distribution can be approximated by
a unimodal Gaussian distribution with possibly large variance.
Therefore, low-variance out-of-distribution data
with mean identical to in-distribution data can contain
completely different images.

Our analysis indicates that
the squared distance from the mean of each mode is an important factor
of likelihood assignment.
We later show that our experimental results are consistent with this analysis.
Note that
this analysis does not provide an exhaustive explanation for our results,
as our experiments show that the squared distance
is not the only important factor underlying the likelihood assigned to out-of-distribution inputs
(Section~\ref{subsection:twolabels}).
However, our simple analysis provides an intuitive interpretation of our experimental results
similar to the suggestion from the analysis by Nalisnick \etal \cite{Nalisnick2019}.

\section{Proposed Model}
We replace the prior distributions of
deep generative models with mixture distributions $\sum_{i=1}^K p_i/K$
that are not trainable,
and we assume that all components are uniformly weighted.
Although some previous studies have performed clustering using VAEs
with a trainable multimodal prior distribution
\cite{Dilokthanakul_Deep_2016, Tomczak_VAE_2017},
we manually assign each input to
a component of the prior distribution before training.
We simply use the labels of the data sets
or apply k-means clustering on the training data
to decide on which components to assign to.
During training,
the likelihood for each input is evaluated using
a different unimodal prior distribution $p_i$
(using a different index $i$ for each input),
which is the component of the multimodal prior distribution assigned to each input.
The test likelihood is evaluated on a mixture prior distribution
$\sum_{i=1}^K p_i/K$
without the component assignment used during training.

\begin{figure}[tb]
\begin{center}
\includegraphics[width=0.8\linewidth]{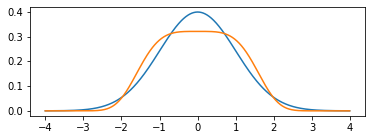}
\end{center}
\caption{Probability density functions of
a standard Gaussian distribution (blue) and a generalized Gaussian distribution
with parameters $\alpha=\sqrt{\Gamma(1/\beta) / \Gamma(3/\beta)}, \beta=4$ (orange).}
\label{fig:gen_gaussian}
\end{figure}

We use Gaussian distributions and generalized Gaussian distributions
for the components of the mixture distributions.
The probability density function of a univariate generalized Gaussian distribution is
\begin{equation}
    p(x; \alpha, \beta) = \frac{\beta}{2\alpha \Gamma(1/\beta)}
    \exp\left(-\left(\frac{|x-\mu|}{\alpha} \right)^{\beta} \right)
\end{equation}
where $\Gamma$ is the Gamma function
and $\alpha, \beta \in (0, +\infty)$ are parameters.
The assumption made in our analysis in Section~\ref{subsection:secondorder}
suggests that the components of prior distributions
should be far from each other and have a small overlap.
We observe that Gaussian distributions are too heavy-tailed
and components must be placed far away from each other
in order to lower the out-of-distribution likelihood.
To deal with this problem,
we propose the use of a generalized Gaussian distribution.
A generalized Gaussian distribution with parameters
$\beta=2$ and $\alpha=\sqrt{2}$ is equal to a standard Gaussian distribution.
A generalized Gaussian distribution with large $\beta$ is light-tailed,
so we use a generalized Gaussian distribution
with parameters $\alpha=\sqrt{\Gamma(1/\beta) / \Gamma(3/\beta)}, \beta=4$
(Section~\ref{subsection:twolabels}).
Note that the variance of a generalized Gaussian distribution
using $\alpha=\sqrt{\Gamma(1/\beta) / \Gamma(3/\beta)}$ is one.

\section{Experiments \label{section:experiments}}
We evaluate the effect of a multimodal prior distribution
on the likelihoods assigned to out-of-distribution inputs
on two pairs of real image data sets:
Fashion-MNIST vs.\ MNIST, and CIFAR-10 vs.\ SVHN.

\subsection{Data Sets}
We use two pairs of image data sets.
The first pair is
Fashion-MNIST \cite{Xiao2017Fashion-MNIST:Algorithms} (training data)
and MNIST \cite{LeCun1998Gradient-BasedRecognition} (out-of-distribution inputs).
The second pair is
CIFAR-10 \cite{Krizhevsky2009LearningImages} (training data)
and SVHN \cite{Netzer2011ReadingLearning} (out-of-distribution inputs).
For training,
we use a small subset of the data sets,
as using all images requires a large number of clusters
to lower the out-of-distribution likelihood.
For CIFAR-10,
10\% random width and height shifting is applied during training
as data augmentation.

\subsection{Model Architecture and Training Details}
Our implementation of VAE is based on the architecture
described in \cite{Rosca2018DistributionInference, Nalisnick2019}.
Both the encoder and the decoder are convolutional neural networks.
Our implementation of Glow is based on the authors' code hosted at
OpenAI's open source repository\footnote{\url{https://github.com/openai/glow}}.
To remove spatial dependencies on the latent variables,
we do not use the multi-scale architecture,
and apply $1\times 1$ convolution over three dimensions
(width, height, channel) after the decoder.
Further details are discussed in Appendix~\ref{settings}.

\subsection{Two Labels and Two Modes \label{subsection:twolabels}}
\begin{figure}[tb]
\begin{center}
\begin{subfigure}[t]{\linewidth}
    \center
    \includegraphics[width=0.9\textwidth]{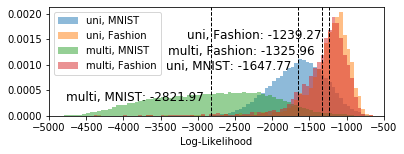}
    \caption{VAE \label{fig:vae}}
\end{subfigure}
\begin{subfigure}[t]{\linewidth}
    \center
    \includegraphics[width=0.9\textwidth]{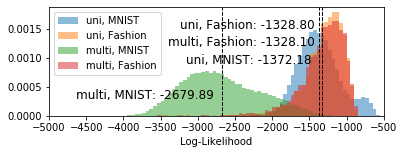}
    \caption{Glow \label{fig:glow_train}}
\end{subfigure}
\end{center}
\caption{
Histograms of the log-likelihoods
assigned by VAEs and Glow trained on Fashion-MNIST (label 1 and 7).
``uni'' denotes a standard Gaussian prior
and ``multi'' denotes a bimodal Gaussian mixture prior.
For Fashion-MNIST, we report likelihoods evaluated on test data.
Bimodal priors mitigate the out-of-distribution problem.}
\label{fig:2labels}
\end{figure}

We first analyze our model on simple data sets of images.
Here, models are trained on images in
label 1 (Trouser) and 7 (Sneaker) in Fashion-MNIST.
We compare two types of prior distributions: a standard Gaussian distribution and
a bimodal Gaussian mixture distribution.
The means of the bimodal priors are $[\pm 75, 0, \ldots, 0]$ for VAE
and $[\pm 50, 0, \ldots, 0]$ for Glow.
The variances are $\mathrm{diag}([1, \ldots, 1])$ on all the components.
Figure~\ref{fig:2labels} shows that
the models using multimodal prior distributions correctly
assign low likelihood to MNIST, the out-of-distribution data,
while the models using unimodal prior distributions assign
high likelihood to MNIST.

\begin{table}[tb]
\center
\begin{tabular}{|c|c|l|r|}
\hline
\multicolumn{1}{|c|}{Model} & \multicolumn{1}{|c|}{Prior} &
\multicolumn{1}{|c|}{Training Data} & \multicolumn{1}{|c|}{NLL} \\
\hline \hline
VAE   & uni   & Fashion-MNIST (1, 7) & 1647.76 \\
VAE   & uni   & Fashion-MNIST (1)    & 1976.73 \\
VAE   & uni   & Fashion-MNIST (7)    & 1578.43 \\
VAE   & multi & Fashion-MNIST (1, 7) & 2821.97 \\
\hline
Glow   & uni   & Fashion-MNIST (1, 7) & 1372.18 \\
Glow   & uni   & Fashion-MNIST (1)    & 1791.96 \\
Glow   & uni   & Fashion-MNIST (7)    & 1595.10 \\
Glow   & multi & Fashion-MNIST (1, 7) & 2679.89 \\
\hline
\end{tabular}
\caption{Negative log-likelihoods assigned to MNIST by the models trained on
Fashion-MNIST.
Fashion-MNIST ($i$) indicates that the model is trained on the images in the $i$-th label.
``uni'' denotes a standard Gaussian prior
and ``multi'' denotes a bimodal Gaussian mixture prior.
The unimodal prior models
trained only on images for one label of Fashion-MNIST
still exhibit the out-of-distribution phenomenon for MNIST
when compared to multimodal prior models.}
\label{tab:unimodal}
\end{table}

\paragraph{Multi-Modal Priors Force Out Out-of-Distribution Points}
\begin{figure}[t]
\begin{center}
\begin{subfigure}[t]{\linewidth}
    \center
    \includegraphics[width=0.8\linewidth]{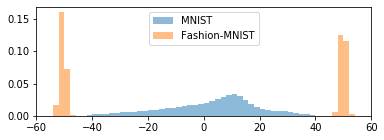}
    \caption{bimodal, label 1 and 7}
\end{subfigure}
\begin{subfigure}[t]{\linewidth}
    \center
    \includegraphics[width=0.8\linewidth]{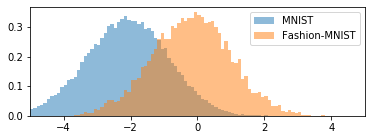}
    \caption{Unimodal, label 7}
\end{subfigure}
\end{center}
\caption{Histograms of the first-idx of the latent variables on Glow
trained on label 1 and 7
and the 613rd-idx trained only on label 7 of Fashion-MNIST.
For the unimodal prior model,
we select the dimension of the latent variable with the largest absolute mean
for MNIST.
Further results are reported in Appendix~\ref{appendix:latentvariables}.}
\label{fig:lv_forceout}
\end{figure}

Our analysis in Section~\ref{subsection:secondorder} suggests that
multimodal prior models mitigate the out-of-distribution problem
because each component is trained on simpler data.
However, although the complexity of data allocated to 
each component is important,
unimodal prior models 
trained on data allocated to a single component
still assign high likelihoods to out-of-distribution inputs
when compared to multimodal prior models (Table~\ref{tab:unimodal}).
Figure~\ref{fig:lv_forceout} shows that the model with the multimodal prior
correctly places MNIST in an out-of-distribution area
wihin the latent variable space.
In contrast, MNIST and Fashion-MNIST (label 7) have large overlap
within the latent variable space of
the model with a unimodal prior trained only on label 7 of Fashion-MNIST.
These results imply that separating in-distribution data
in the latent variable space by using a multimodal prior distribution
has a strong effect of forcing out-of-distribution points out of
high-likelihood areas.
This observation suggests a new approach for mitigating the out-of-distribution phenomenon;
improving latent variable design.

\paragraph{Distance between Two Components}
\begin{figure}[tb]
\begin{center}
    \center
    \includegraphics[width=0.9\linewidth]{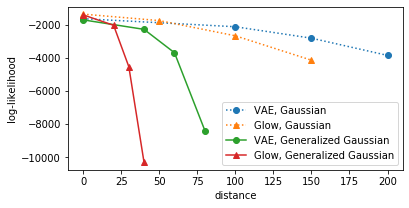}
\end{center}
\caption{Relationships of the distance between two components
and the mean log-likelihoods assigned to
MNIST by models trained on Fashion-MNIST (label 1 and 7).
While likelihoods assigned to out-of-distribution inputs are
sensitive to the distance between components
regardless of component choice,
the Gaussian mixture priors require much larger distances
to lower the likelihood assigned to out-of-distribution inputs.
The histograms and the mean values of the log-likelihoods
are reported in Appendix~\ref{appendix:distance}}
\label{fig:distance-graph}
\end{figure}

We analyze the relationship of
the likelihoods assigned to out-of-distribution inputs
and the distance between two components
using two types of distributions:
Gaussian and generalized Gaussian mixture distributions.
Figure~\ref{fig:distance-graph} shows the mean log-likelihoods assigned to MNIST
by the models trained on Fashion-MNIST (label 1 and 7)
with different types and component distances of prior distributions.
Likelihoods assigned to out-of-distribution inputs are
sensitive to the distance between components
regardless of the component choice.
However, models using Gaussian mixture priors require
larger distances to lower the out-of-distribution likelihoods.
A generalized Gaussian prior ($\beta=4$) is particularly
effective at assigning low likelihoods to out-of-distribution inputs
even with much smaller distances between components.
The means of the bimodal distributions are $[\pm d/2, 0, \ldots, 0]$,
and the variance is $\mathrm{diag}([1, \ldots, 1])$ for all the components.
Note that the likelihoods assigned to the test data of Fashion-MNIST 
(label 1 and 7) are relatively
unaffected by the distance between two components
(Appendix \ref{appendix:distance}).

\paragraph{Second Order Analysis \label{subsection:experiment-soa}}
\begin{figure}[t]
\begin{center}
\begin{subfigure}[t]{\linewidth}
    \center
    \includegraphics[width=0.8\linewidth]{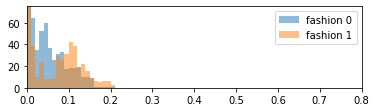}
    \caption{Per-dimensional empirical mean of the squared distance between the mean image
        of each component of the VAE trained on Fashion-MNIST (label 1 and 7)
        and images in Fashion-MNIST (label 1 and 7). \label{fig:variance_fashion}}
\end{subfigure}
\begin{subfigure}[t]{\linewidth}
    \center
    \includegraphics[width=0.8\linewidth]{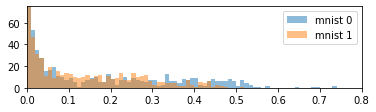}
    \caption{Per-dimensional empirical mean of the squared distance between the mean image
        of each component of the VAE trained on Fashion-MNIST (label 1 and 7)
        and images in MNIST. \label{fig:variance_mnist}}
\end{subfigure}
\begin{subfigure}[t]{\linewidth}
    \center
    \includegraphics[width=0.8\linewidth]{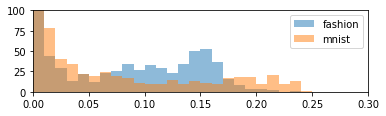}
    \caption{Per-dimensional variance of images in Fashion-MNIST (label 1 and 7)
    and MNIST\label{fig:variance_full}}
\end{subfigure}
\end{center}
\caption{
Comparison of the experimental results with
the suggestion based on the analysis in Section~\ref{subsection:secondorder}.
(\subref{fig:variance_fashion}), (\subref{fig:variance_mnist})
The per-dimensional empirical mean of the squared distance from the mean image of each component
of the VAE with a bimodal prior distribution.
The model is trained on Fasion-MNIST (label 1 and 7)
and test images in Fashion-MNIST (label 1 and 7) and MNIST are assumed to be allocated to the nearest component
in the latent variable space.
``fashion $i$'' and ``mnist $i$'' denote the data allocated to the $i$-th component.
(\subref{fig:variance_full}) The per-dimensional variances over pixels
of images in MNIST and Fashion-MNIST.
The y-axis is clipped for visualization.}
\label{fig:variance}
\end{figure}

Our analysis in Section~\ref{subsection:secondorder} suggests that
the conditional expectation of the squared distances
from the mean of the images generated from the modes, \ie
$\sigma^2_{\mathcal D^*_i, h, w, c}, \sigma^2_{\mathcal D_i, h, w, c}$,
influence the assigned likelihoods.
We show that our experimental results are consistent with the analysis.
Note that this is not the only influencing factor that
lowers the out-of-distribution likelihood as mentioned above.
Figure~\ref{fig:variance_fashion}, \subref{fig:variance_mnist}
show histograms of the per-dimensional empirical mean of the squared distance from
the mean of the images generated from each component of the VAE with a bimodal prior distribution,
which are the empirical values of
$\sigma^2_{\mathcal D^*_i, h, w, c}$ and $\sigma^2_{\mathcal D_i, h, w, c}$,
respectively.
The pixel values are scaled to $[0, 1]$.
As explained in Section~\ref{subsection:secondorder},
we consider each image as allocated to
the closest component in the latent variable space,
and the squared distance from the mean is calculated over the pixel space.
These results are evaluated on the VAE using a bimodal Gaussian mixture prior
with means $[\pm 75, 0, \ldots, 0]$ trained on Fashion-MNIST (label 1 and 7).
Figure~\ref{fig:variance_fashion}, \subref{fig:variance_mnist} show that 
Fashion-MNIST has lower empirical values of $\sigma^2_{\mathcal D^*_i, h, w, c}$ compared to
those of MNIST $\sigma^2_{\mathcal D_i, h, w, c}$.
The results are consistent with the implication of our analysis
that higher likelihood is assigned if the values are small.

Figure~\ref{fig:variance_full} shows the per-dimensional variance
of images in Fashion-MNIST (label 1 and 7) and MNIST.
In contrast to our analysis,
the analysis by Nalisnick \etal \cite{Nalisnick2019}
for unimodal prior models suggests that 
the models assign higher likelihood to an adversarial distribution
if the per-dimensional variance is small.
As has been reported in \cite{Nalisnick2019},
most pixels of images found in MNIST have low variance,
and this is consistent with the result that MNIST is assigned higher likelihood
when a standard Gaussian prior is used.
The differences between these two types of histograms provide
an intuitive explanation for the difference between
likelihoods assigned to out-of-distribution inputs by models using
unimodal and multimodal prior distributions.

\subsection{Results on Complex Data Sets \label{subsection:complexdatasets}}
We evaluate our proposition on more complex data sets.
While multimodal priors assign lower likelihoods,
the effect is especially limited on Glow.
The results on Glow may be affected by the spatial dependencies on the latent variables,
and our efforts to remove the dependencies may not be sufficient.
Our observations suggest that Glow requires further modifications to solve this problem,
so we leave this problem for future work.
More investigation into latent variable space as well as separation of data sets
is required for further performance.
Alternatively,
our method can be used in tandem with other techniques
such as \cite{Hendrycks2019DeepExposure, Nalisnick2019DetectingTypicality}.

\paragraph{Fashion-MNIST (label 0, 1, 7, 8) vs MNIST}
\begin{figure}[t]
\begin{subfigure}[t]{\linewidth}
    \center
    \includegraphics[width=0.9\linewidth]{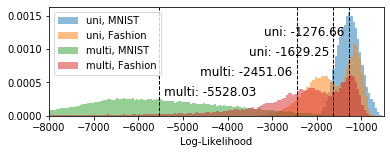}
    \caption{VAE}
\end{subfigure}
\begin{subfigure}[t]{\linewidth}
    \center
    \includegraphics[width=0.9\linewidth]{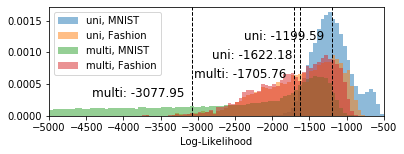}
    \caption{Glow}
\end{subfigure}
\caption{Likelihoods assigned by the models trained on Fashion-MNIST
(label 0, 1, 7, 8).
The models with unimodal priors assign higher likelihood to MNIST,
whereas those using multimodal priors mitigate this problem.
}

\label{fig:fashion_4labels}
\end{figure}

We evaluate our method on a VAE trained on label 0, 1, 7, and 8.
(T-shirt, Trouser, Sneaker, and Bag).
We compare a standard Gaussian prior and a quad-modal Gaussian mixture prior.
For the VAE,
the means are $[150, 0, \ldots, 0]$, $[0, 150, 0 \ldots, 0]$,
$[0, 0, 150, 0, \ldots, 0]$, and $[0, 0, 0, 150, 0, \ldots, 0]$).
However, we observe that this scheme
of placing modes on differing dimensions does not work on Glow.
Hence, we use $[200\times i, 0, \ldots, 0] (i=0, 1, 2, 3)$ for Glow.
The variances are $\mathrm{diag}([1, \ldots, 1])$ for all components.
Figure~\ref{fig:fashion_4labels} shows that
the models using standard Gaussian priors produce the out-of-distribution phenomenon on MNIST,
while the models using Gaussian mixture priors do to a lesser degree.

\paragraph{CIFAR-10 (label 0 and 4) vs SVHN}
\begin{figure}[t]
\begin{center}
\begin{subfigure}[t]{\linewidth}
    \center
    \includegraphics[width=0.9\linewidth]{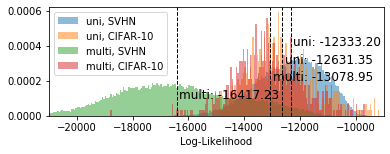}
    \caption{VAE}
\end{subfigure}
\begin{subfigure}[t]{\linewidth}
    \center
    \includegraphics[width=0.9\linewidth]{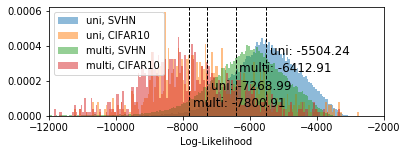}
    \caption{Glow}
\end{subfigure}
\end{center}
\caption{Likelihood assigned by models trained on
CIFAR-10 (label 0 and 7).
The models using standard Gaussian priors assign higher likelihood to SVHN.
The models using multimodal priors mitigate this problem
while the effect is limited on Glow.}
\label{fig:CIFAR-10_results}
\end{figure}

We find that images for one label in CIFAR-10 are still too diverse for our method.
Therefore, we apply k-means and separate the images in label 0 and 4 of
CIFAR-10 into four respective clusters,
and we use images in one cluster of each label.
We compare the models using a standard Gaussian prior and
a bimodal Gaussian mixture prior
with means $[\pm 100, 0, \ldots, 0]$ for VAE,
and $[\pm 200, 0, \ldots, 0]$ for Glow.
Figure~\ref{fig:CIFAR-10_results} shows
that the models with multimodal priors
assign lower likelihoods to SVHN
compared to the models using unimodal prior distributions.
However, this effect is limited on Glow.
We hypothesize that images in each cluster are still too diverse
for Glow with our settings.
These results imply that it is difficult to adopt our method
if a data set does not consist of low-variance and distant clusters,
and further study is required particularly for Glow.

\section{Conclusion and Discussion}
We analyzed the influence of prior distribution choice of deep generative models
on the likelihoods assigned to out-of-distribution inputs.
Recent work \cite{Nalisnick2019, Choi2019WAICDetection}
on deep generative models with unimodal prior distributions
has shown that these models can assign higher likelihoods
to out-of-distribution inputs than to training data.
In this paper,
we showed that models using multi-modal prior distributions lower
the likelihoods assigned to out-of-distribution inputs
for Fashion-MNIST vs.\ MNIST and CIFAR10 vs.\ SVHN.
We also provided theoretical explanations for the advantages of
the use of multi-modal prior distributions.

Unfortunately,
our experimental results suggested that it is difficult to apply our method
to complex data sets even when we use prior knowledge.
Thus, our work demonstrates the limitation of the high-dimensional likelihoods yet again,
and encourages future work on alternative metrics such as
\cite{Choi2019WAICDetection, Nalisnick2019DetectingTypicality}.
Nevertheless, our work is the first to show that
likelihoods assigned to out-of-distribution inputs are
affected by the choice of the prior distribution,
which has been mainly studied as a way to improve the representative ability of
deep generative models for in-distribution data.
Our observations motivate further study on
the prior distributions of deep generative models,
as well as on methods to control the structure of the latent variables
to make the model likelihood sensitive to out-of-distribution inputs.

\ifcvprfinal
\section*{Acknowledgements}
This paper has benefited from advice and English language editing from
Masayuki Takeda.
This work was supported by JSPS KAKENHI (JP19K03642, JP19K00912) and RIKEN AIP Japan.

\fi

{\small
\bibliographystyle{ieee_fullname}
\bibliography{egbib}
}

\clearpage

\begin{center}
{\large \bf 
Supplimentary Material of \\
``Likelihood Assignment for Out-of-Distribution Inputs in 
Deep Generative Models is Sensitive to Prior Distribution Choice''
\par}
\end{center}
\vspace{\baselineskip}

\appendix
\section{Second Order Analysis \label{appendix:secondorder}}
In this section,
we present detailed explanations for the analysis in Section~\ref{subsection:secondorder}.
We adopt the assumption that the probability distribution function of
of the given generative model $p$ can be approximated by mixture distribution
$\log p(\mathbf x; \theta) \simeq \log \frac{1}{K}\sum_{i=1}^K p_i(\mathbf x; \theta)$,
where $p_i$ corresponds to each component that can be approximated by a Gaussian distribution.
For simplicity,
we assume that the components are assigned the uniform weights,
and have equal variances.
In addition, corresponding to the nature of the data sets that we are considering,
we assume that components of the distribution are
far from each other and have small variances.
Under the assumptions, we can approximate the probability distribution for each input
by taking the value from the component that yields the maximum value for the data:
$\log p(\mathbf x; \theta) \simeq \log \frac{1}{K}\sum_{i=1}^K p_i(\mathbf x; \theta)
\simeq \max_i \log \frac{1}{K} p_i(\mathbf x; \theta)$.
Thus, we can write the expectation by the training data distribution $p^*$ as
\begin{align}
    \mathbb E_{p^*}[\log p(\mathbf x; \theta)]
    &= \sum_{i=1}^K w^*_i \mathbb E [\log p(\mathbf x; \theta) | \mathcal D^*_i] \nonumber \\
    &\simeq \sum_{i=1}^K w^*_i \mathbb E \left[\left. \log \frac{1}{K}\sum_{j=1}^K p_j(\mathbf x; \theta)
        \right| \mathcal D^*_i \right] \nonumber \\
    &\simeq \sum_{i=1}^K w^*_i \mathbb E [\log p_i(\mathbf x; \theta) | \mathcal D^*_i] - \log K
\end{align}
where $\mathcal D^*_i$ represents in-distribution data allocated to the $i$-th component
and $w^*_i$ is the ratio of data allocated to the $i$-th component
satisfying $\sum_{i=1}^K w^*_i = 1$.
We can also expand the expectation by the adversarial distribution $q$ as
$\mathbb E_{q}[\log p(\mathbf x; \theta)] \simeq
\sum_{i=1}^K w_i \mathbb E [\log p_i(\mathbf x; \theta) | \mathcal D_i] - \log K$
where $\mathcal D^*_i$ represents out-of-distribution data allocated to the $i$-th component,
and $w_i$ is the ratio of data allocated to $i$-th component
satisfying $\sum_{i=1}^K w_i = 1$.

Since we assume that each component can be approximated by a Gaussian distribution,
we use second order approximation for each component:
$\log p_i(\mathbf x; \theta) \simeq \log p_i(\mathbf{\bar{x}}_i; \theta)
+ \nabla_{\mathbf{\bar{x}}_i} \log p_i(\mathbf{\bar{x}}_i; \theta)^T(\mathbf x-\mathbf{\bar{x}}_i)
+ \frac{1}{2} \text{Tr} \{\nabla^2_{\mathbf{\bar{x}}_i} \log p_i(\mathbf{\bar{x}}_i; \theta)
(\mathbf x-\mathbf{\bar{x}}_i)(\mathbf x-\mathbf{\bar{x}}_i)^T \}$.
Here, $\mathbf{\bar{x}}_i$ is the mean of
images generated from each component.
Therefore, $\nabla_{\mathbf{\bar{x}}_i} \log p(\mathbf{\bar{x}}_i; \theta)^T(\mathbf x-\mathbf{\bar{x}}_i) \simeq 0$
since $\mathop{\rm argmax}_x \log p_i(\mathbf x; \theta) \simeq \mathbf{\bar{x}}_i$.
Thus we can expand the conditional expectation as
\begin{align}
    \mathbb E&[\log p_i(\mathbf x; \theta) | \mathcal D^*_i] \nonumber \\
    &\simeq \mathbb E \bigg[\log p_i(\mathbf{\bar{x}}_i; \theta) \nonumber \\
    &\hspace{1em} + \left. \frac{1}{2} \text{Tr} \{\nabla^2_{\mathbf{\bar{x}}_i}
        \log p_i(\mathbf{\bar{x}}_i; \theta) (\mathbf x-\mathbf{\bar{x}}_i)(\mathbf x-\mathbf{\bar{x}}_i)^T \} 
        \right| \mathcal D^*_i \bigg] \nonumber \\
    &= \log p_i(\mathbf{\bar{x}}_i; \theta)
        + \frac{1}{2} \text{Tr} \{\nabla^2_{\mathbf{\bar{x}}_i}
        \log p_i(\mathbf{\bar{x}}_i; \theta) \mathbf \Sigma_{\mathcal D^*_i} \}
\end{align}
where $\mathbf \Sigma_{\mathcal D^*_i} = \mathbb E[(\mathbf x-\mathbf{\bar{x}}_i)(\mathbf x-\mathbf{\bar{x}}_i)^T | \mathcal D^*_i]$,
and it is assumed to be diagonal as in \cite{Nalisnick2019}.
Furthermore, $\mathbb E[\log p_i(\mathbf x; \theta) | \mathcal D_i] = 
\log p_i(\mathbf{\bar{x}}_i; \theta) + \frac{1}{2} \text{Tr} \{\nabla^2_{\mathbf{\bar{x}}_i}
\log p_i(\mathbf{\bar{x}}_i; \theta) \mathbf \Sigma_{\mathcal D_i} \}$
where $\mathbf \Sigma_{\mathcal D_i} = \mathbb E[(\mathbf x-\mathbf{\bar{x}}_i)(\mathbf x-\mathbf{\bar{x}}_i)^T | \mathcal D_i]$.
Note that $\mathbf \Sigma_{\mathcal D^*_i}$ and  $\mathbf \Sigma_{\mathcal D_i}$
are not the variance matrices as
$\mathbf{\bar{x}}_i$ are the mean images of the generative model.

Because we assume that variances of all components are identical,
$\log p_i(\mathbf{\bar{x}}_i; \theta)$ can be approximated to be identical for all $i$.
Finally,
we can write the difference of the two log-likelihoods (Equation~\ref{eq:diff_of_likelihoods}) in
a relatively simple form in parallel with
the first line of Equation~5 in \cite{Nalisnick2019}:
\begin{align} \label{equation:intermediate_form}
    & \mathbb E_{q}[\log p(\mathbf x; \theta)] - \mathbb E_{p^*}[\log p(\mathbf x; \theta)] 
    \nonumber \\
    &\simeq \sum_{i=1}^K w_i \mathbb E [\log p_i(\mathbf x; \theta) | \mathcal D_i]
    \nonumber \\
    &\hspace{1em} - \sum_{i=1}^K w^*_i \mathbb E [\log p_i(\mathbf x; \theta) | \mathcal D^*_i]
    \nonumber \\
    &= \frac{1}{2} \text{Tr} \left\{
        \sum_{i=1}^K w_i \nabla^2_{\mathbf{\bar{x}}_i}
        \log p_i(\mathbf{\bar{x}}_i; \theta) \mathbf \Sigma_{\mathcal D_i}  \right.
    \nonumber \\
    &\hspace{4em} \left.
        -\sum_{i=1}^K w^*_i \nabla^2_{\mathbf{\bar{x}}_i}
        \log p_i(\mathbf{\bar{x}}_i; \theta) \mathbf \Sigma_{\mathcal D^*_i} \right\}.
\end{align}

If we assume that each $p_i$ is precisely a Gaussian distribution,
we can simply compute the Hessian in Equation~\ref{equation:intermediate_form}.
However, because this assumption is too strong,
Nalisnick \etal \cite{Nalisnick2019} expanded this formula by
adopting the assumption that the generative model is constant-volume Glow (CV-Glow).
Although we do not use CV-Glow in our experiments,
we apply the expression derived by Nalisnick \etal \cite{Nalisnick2019}:
\begin{align}
    &\text{Tr} \left\{
        \nabla^2_{\mathbf{\bar{x}}_i} \log p_i(\mathbf{\bar{x}}_i; \theta) \mathbf \Sigma_{\mathcal D_i}  \right\}
    \nonumber \\
    &= -\frac{1}{2\sigma^2_{\psi}}
        \sum_{c=1}^C \left(\prod_{l=1}^L \sum_{j=1}^{C_l} u_{l, c, j} \right)^2
        \sum_{h, w} \sigma^2_{\mathcal D_i, h, w, c}
\end{align}
where $\sigma^2_{\psi}$ is the variance of a component of the prior distribution
(we assume all components have identical variance)
and $\sigma^2_{\mathcal D_i, h, w, c}$ are
diagonal elements of $\Sigma_{\mathcal D_i}$.
$u_{l, c, j}$ is the weight of the $l$-th 1x1 convolution
of Glow, which is fixed for any inputs.
$h$ and $w$ index the input spatial dimensions,
$c$ indexes the input channel dimensions,
$l$ indexes the series of flows,
and $j$ indexes the column dimensions of the $C_l \times C_l$ kernel.
We assume that each component of the generative model $p_i(\mathbf x; \theta)$
corresponds to a component of prior distribution $p_i(\mathbf z; \psi)$.
Finally, we arrive Equation~\ref{eq:final_formula}.

\section{Experimental Settings \label{settings}}
We present model architectures and training settings
of the experiments shown in Section~\ref{section:experiments}.

\begin{table}[b]
\begin{center}
\begin{subtable}{\linewidth}\centering
\begin{tabular}{|l|c|c|c|c|}
\hline
Operation & Kernel & Strides & Channels & Pad \\
\hline\hline
Convolution & $5\times 5$ & 2 & 8 & 1 \\
Convolution & $5\times 5$ & 1 & 16 & 1 \\
Convolution & $5\times 5$ & 2 & 32 & 1 \\
Convolution & $5\times 5$ & 1 & 64 & 1 \\
Convolution & $5\times 5$ & 2 & 64 & 1 \\
Fully-Connected & --- & --- & $50\times 2$ & --- \\
\hline
\end{tabular}
\caption{Encoder. The outputs are means (50) and log variances (50).}
\end{subtable}
\vspace{1em}

\begin{subtable}{\linewidth}\centering
\begin{tabular}{|l|c|c|c|c|}
\hline
Operation & Kernel & Strides & Channels & Pad \\
\hline\hline
Fully-Connected & --- & --- & 3136 & --- \\
Reshape & --- & --- & 64 & --- \\
Convolution & $5\times 5$ & 2 & 64 & 2 \\
Convolution & $5\times 5$ & 2 & 32 & 1 \\
Convolution & $5\times 5$ & 1 & 64 & 1 \\
Convolution & $4\times 4$ & 1 & 256 & 1 \\
\hline
\end{tabular}
\caption{Decoder. ``Convolution'' in the decoder is transposed convolution.
The reshape operation reshape
the latent variables sized $3,136$ to $7 \times 7 \times 64$.}
\end{subtable}

\end{center}
\caption{Model architecture of VAE for Fashion-MNIST and MNIST.
``Channels'' denotes the size of the output channel,
and ``Pad'' denotes paddings.
\label{table:vae_structure}}
\end{table}

\begin{table}[b]
\begin{center}
\begin{subtable}{\linewidth}\centering
\begin{tabular}{|l|c|c|c|c|}
\hline
Operation & Kernel & Strides & Channels & Pad \\
\hline\hline
Convolution & $5\times 5$ & 2 & 8 & 1 \\
Convolution & $5\times 5$ & 1 & 16 & 1 \\
Convolution & $5\times 5$ & 2 & 32 & 1 \\
Convolution & $5\times 5$ & 1 & 64 & 1 \\
Convolution & $5\times 5$ & 2 & 64 & 1 \\
Fully-Connected & --- & --- & $100\times 2$ & --- \\
\hline
\end{tabular}
\caption{Encoder. The outputs are means (50) and log variances (50).}
\end{subtable}
\vspace{1em}

\begin{subtable}{\linewidth}\centering
\begin{tabular}{|l|c|c|c|c|}
\hline
Operation & Kernel & Strides & Channels & Pad \\
\hline\hline
Convolution & $4\times 4$ & 1 & 64 & 0 \\
Convolution & $4\times 4$ & 2 & 32 & 1 \\
Convolution & $4\times 4$ & 2 & 32 & 1 \\
Convolution & $4\times 4$ & 2 & $256\times3$ & 1 \\
\hline
\end{tabular}
\caption{Decoder. ``Convolution'' in the decoder is transposed convolution.
The reshape operation reshapes
the latent variables sized $3,136$ to $7 \times 7 \times 64$.}
\end{subtable}

\end{center}
\caption{Model Architecture of VAE for CIFAR-10 and SVHN.
``Channels'' is the size of output channel,
and ``Pad'' is paddings.
\label{table:vae_structure_cifar}}
\end{table}

\paragraph{VAE}
Our implementation of VAE \cite{Kingma_Auto_2013} is based on the architecture
described in \cite{Rosca2018DistributionInference, Nalisnick2019}.
Both the encoder and the decoder are convolutional neural networks
described in Table~\ref{table:vae_structure} and \ref{table:vae_structure_cifar}.
We use batch normalization \cite{Ioffe2015BatchShift}
after every convolutional layer except for the last layer
of the encoder and the decoder.
All the convolutional layers in the decoder use ReLU \cite{Nair2010RectifiedMachines} 
as an activation function after batch normalization.
After the final layer of the decoder, we apply the softmax function,
and assume i.i.d.\ categorical distributions on pixels
as visual distributions.

We perform training for 1,000 epochs using the Adam optimizer \cite{Kingma_Adam_2014}
($\beta_1=0.5, \beta_2=0.9$) with a constant learning rate of $1\mathrm{e}{-3}$.
We use 5,000 sample points to approximate test likelihoods.

\paragraph{Glow}
For our experiments of Glow \cite{Kingma2018Glow:Convolutions},
our implementation is based on the code hosted at
OpenAI's open source repository\footnote{\url{https://github.com/openai/glow}}.
For Fashion-MNIST vs.\ MNIST, we use 1 block of 32 affine coupling layers,
squeezing the spatial dimension after the 16-th layer.
For CIFAR-10 vs.\ SVHN, we use 1 block of 24 affine coupling layers,
squeezing the spatial dimension after the 8-th and 16-th layer.

To alleviate the spatial dependencies on the latent variables,
we do not use the multi-scale architecture,
which splits the latent variables after squeezing
\cite{Dinh2017DensityNVP}.
In addition,
we apply $1\times 1$ convolution over three dimensions
(width, height, channel) after the encoder,
and apply the inverse operation before the decoder.
In the implementation,
we add the code in Listing~\ref{lst:permutation} after the encoder,
and add the inverse operation before the decoder.
Moreover, we add a small positive value (0.1 in our implementation)
to the scale of affine coupling layers 
to stabilize the training as suggested at
\footnote{\url{https://github.com/openai/glow/issues/40}}.
While Nalisnick \etal \cite{Nalisnick2019} remove actnorm
and apply their original initialization scheme,
we use actnorm and apply the original initialization scheme 
in the OpenAI's code.

We perform training for 1,000 epochs using the Adam optimizer
in accordance with the OpenAI's code.
We use a learning rate of $1\mathrm{e}{-3}$,
which is linearly annealed from zero over the first 10 epochs.

\begin{lstlisting}[language=Python, float=tp, numbers=left,
linewidth=\linewidth,
captionpos=b, caption={Code added for permutation after the encoder of Glow
to remove the spacial dependencies on latent variables.
The inverse operation is added before the decoder.},
label={lst:permutation},
breaklines=true]
z = tf.transpose(z, perm=[0, 3, 2, 1])
z, logdet = invertible_1x1_conv("invconv1", z, logdet)
z = tf.transpose(z, perm=[0, 3, 2, 1])
z = tf.transpose(z, perm=[0, 1, 3, 2])
z, logdet = invertible_1x1_conv("invconv2", z, logdet)
z = tf.transpose(z, perm=[0, 1, 3, 2])
z, logdet = invertible_1x1_conv("invconv3", z, logdet)
\end{lstlisting}

\section{Simple Artificial Data \label{appendix:artificial}}
\begin{figure}[tb]
\begin{subfigure}[t]{\linewidth}
    \center
    \includegraphics[width=\linewidth]{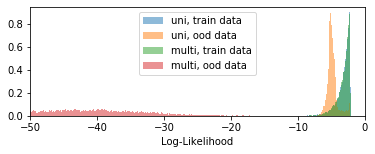}
    \caption{2 dimensional data \label{fig:simpledata_likelihood}}
\end{subfigure}
\begin{subfigure}[t]{\linewidth}
    \center
    \includegraphics[width=\linewidth]{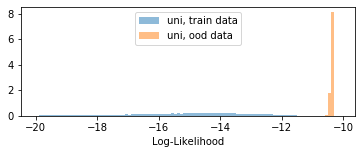}
    \caption{10 dimensional data, Standard Gaussian prior \label{fig:10dstd}}
\end{subfigure}
\begin{subfigure}[t]{\linewidth}
    \center
    \includegraphics[width=\linewidth]{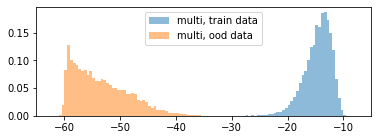}
    \caption{10 dimensional data, multimodal prior \label{fig:10dgmm}}
\end{subfigure}
\caption{(\subref{fig:simpledata_likelihood})
Histograms of the log-likelihoods
assigned to training and out-of-distribution data
by flow-based generative models trained on simple two-dimensional Gaussian mixture data
in Section~\ref{subsection:motivation}.
``uni'' denotes a unimodal prior,
and ``multi'' denotes a multimodal prior.
While a model with a unimodal prior assigns relatively high likelihoods
to out-of-distribution inputs,
a model with a multimodal prior assigns much lower likelihoods
to out-of-distribution inputs.
(\subref{fig:10dstd}, \subref{fig:10dgmm})
The histograms of the log-likelihoods assigned by flow-based generative models
for 10 dimensional data.
The out-of-distribution problem is more serious for high-dimensional data.}
\label{fig:appendix_simpledata}
\end{figure}

For artificial data used in Section~\ref{subsection:motivation},
we compare the likelihoods assigned to in-distribution and out-of-distribution data
to show that a standard Gaussian prior can assign
high likelihoods to out-of-distribution inputs.
The in-distribution data is generated from
a two-dimensional Gaussian mixture distribution whose means are 
$[\pm 3.5, 0]$ and variance is $\mathrm{diag}([0.5, 1])$,
and the out-of-distribution data is sample points from 
a two-dimensional Gaussian distribution with zero mean and 0.01 variance
Figure~\ref{fig:simpledata_visualize}
shows that out-of-distribution inputs does not have any overlap with in-distribution data.
However, Figure~\ref{fig:simpledata_likelihood}
shows that the log-likelihoods assigned for in-distribution and out-of-distribution inputs
by the model using a standard Gaussian prior are similar.
On the contrary,
the model using a multimodal prior distribution assigns much lower likelihoods to
out-of-distribution inputs.

This phenomenon is more serious for high dimensional data.
The in-distribution data is generated from
a 10 dimensional Gaussian mixture distribution whose means are 
$[\pm 3.5, 0, \ldots, 0]$ and variances are $\mathrm{diag}([0.5, 1, \ldots, 1])$
for both components.
The out-of-distribution data is generated from 
a 10 dimensional Gaussian distribution with zero mean and 0.01 variance.
Figure~\ref{fig:10dstd}, \subref{fig:10dgmm}
shows that the log-likelihoods assigned by the model using a standard Gaussian prior
assigned to out-of-distribution inputs
are much higher than those assigned to in-distribution data,
although the model using a multimodal prior assigns much smaller likelihoods
to out-of-distribution inputs.

\section{Mean Images of Clusters}
\begin{figure}[t]
\begin{center}
\includegraphics[width=\linewidth]{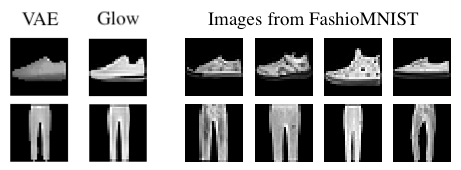}
\end{center}
\caption{Images corresponding to the means of components of
the bimodal prior distributions of VAE and Glow
trained on label 0 and 7 of FahionMNIST (left).
Images in the data set allocated to each cluster (right).}
\label{fig:means_of_clusters}

\begin{center}
\begin{subfigure}{\linewidth}
    \includegraphics[width=\linewidth]{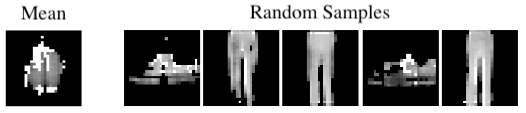}
    \caption{VAE with a unimodal prior}
\end{subfigure}
\begin{subfigure}{\linewidth}
    \includegraphics[width=\linewidth]{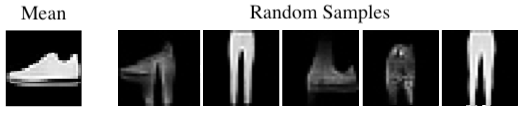}
    \caption{Glow with a unimodal prior}
\end{subfigure}
\end{center}
\caption{Images corresponding to the means of
the unimodal distributions of VAE and Glow
trained on label 0 and 7 of Fashion-MNIST (left),
and images generated from random sampling (right).
The mean image of VAE is dissimilar with training data
while image from random sampling are similar with training data.
While the mean image of Glow is similar to training data,
some images from random sampling of Glow are disimilar with training data.}
\label{fig:std_mean}
\end{figure}

Figure~\ref{fig:means_of_clusters} shows the images corresponding to the means of components of
the bimodal prior distributions of VAE and Glow
trained on label 0 and 7 of Fahion-MNIST.
The means are $[\pm 75, 0, \ldots, 0]$ on the VAE,
and $[\pm 50, 0, \ldots, 0]$ on the Glow.
Figure~\ref{fig:std_mean} shows the images corresponding to the means of
the unimodal prior distributions of the VAE and Glow.
The mean images of the a bimodal prior VAE are similar to
the images in each cluster.
However, the mean image of the standard Gaussian prior VAE is different from
the training data.
For Glow with a standard Gaussian prior,
while the mean image is similar with in-distribution data,
some images from random sampling of Glow
are different from the training data.
The results suggests that the models with unimodal prior distribution
can assign high likelihoods to out-of-distribution inputs
because they can contain out-of-distribution inputs
in their high likelihood areas or typical sets.

\section{K-means Clustering for CIFAR-10}
\begin{figure}[tb]
\begin{subfigure}[t]{\linewidth}
    \center
    \includegraphics[width=\linewidth]{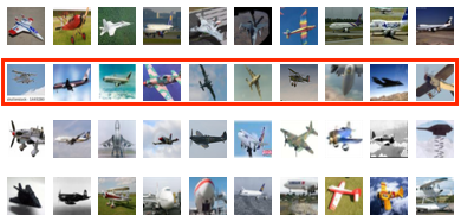}
    \caption{Sample images in four clusters of label 0.
    Each row corresponds to one cluster.
    We use the images in the second row.}
\end{subfigure}
\begin{subfigure}[t]{\linewidth}
    \center
    \includegraphics[width=\linewidth]{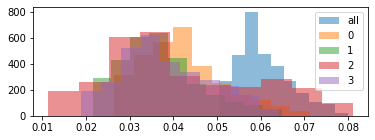}
    \caption{Per-dimensional variance of images in each cluster of label 0.}
\end{subfigure}
\begin{subfigure}[t]{\linewidth}
    \center
    \includegraphics[width=\linewidth]{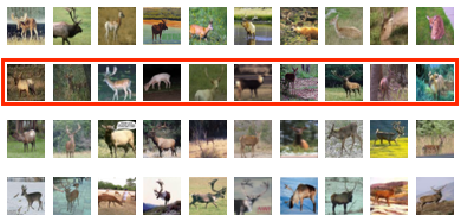}
    \caption{Sample images in four clusters of label 4.
    Each row corresponds to one cluster.
    We use the images in the second row.}
\end{subfigure}
\begin{subfigure}[t]{\linewidth}
    \center
    \includegraphics[width=\linewidth]{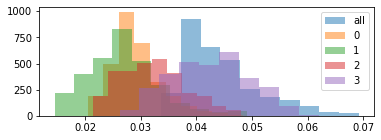}
    \caption{Per-dimensional variance of images in each cluster of label 4.}
\end{subfigure}
\caption{K-means clustering for CIFAR-10 label 0 and 4.}
\label{fig:kmeans_CIFAR-10_stats}
\end{figure}

In the experiments reported in Section~\ref{subsection:complexdatasets},
we separate the images in label 0 and 4 of CIFAR-10 by k-means clustering ($k=4$)
initialized by k-means++ \cite{Arthur2007K-means++:Seeding} respectively.
Figure~\ref{fig:kmeans_CIFAR-10_stats} shows sample images from the clusters
and the per-dimensional variance of the images in each cluster.
The histograms show that k-means clustering successfully decreases
the per-dimensional variance.
In our experiments, we use images in the cluster
corresponding to the second rows.

\section{Additional Experimental Results \label{appendix:experiments}}
We show additional materials for the results reported in 
Section~\ref{section:experiments}.

\subsection{Distance between Two Components \label{appendix:distance}}
\begin{figure}[tb]
\begin{center}
\begin{subfigure}[t]{\linewidth}
    \center
    \includegraphics[width=\linewidth]{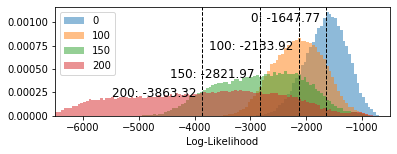}
    \caption{VAE, Gaussian mixture}
\end{subfigure}
\begin{subfigure}[t]{\linewidth}
    \center
    \includegraphics[width=\linewidth]{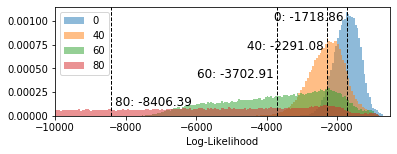}
    \caption{VAE, Generalized Gaussian mixture}
\end{subfigure}
\begin{subfigure}[t]{\linewidth}
    \center
    \includegraphics[width=\linewidth]{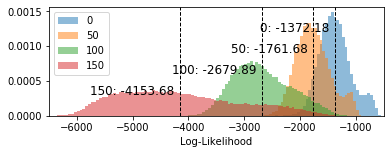}
    \caption{Glow, Gaussian mixture}
\end{subfigure}
\begin{subfigure}[t]{\linewidth}
    \center
    \includegraphics[width=\linewidth]{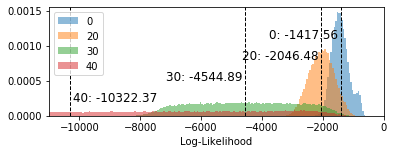}
    \caption{Glow, Generalized Gaussian mixture}
\end{subfigure}
\end{center}
\caption{Distances between two components
and log-likelihoods assigned to
MNIST by models trained on Fashion-MNIST (label 1 and 4).
The results in these images correspond to those in Figure~\ref{fig:distance-graph}.}
\label{fig:appendix-distance-mnist}
\end{figure}

\begin{figure}[t]
\begin{center}
\begin{subfigure}[t]{\linewidth}
    \center
    \includegraphics[width=\linewidth]{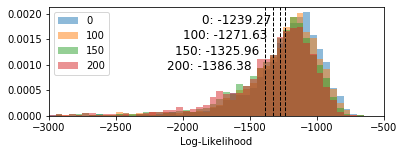}
    \caption{VAE, Gaussian mixture, Fashion-MNIST (1, 7)}
\end{subfigure}
\begin{subfigure}[t]{\linewidth}
    \center
    \includegraphics[width=\linewidth]{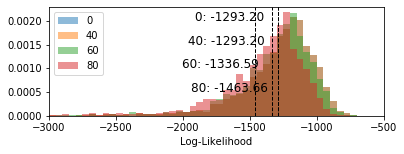}
    \caption{VAE, Generalized Gaussian mixture, Fashion-MNIST (1, 7)}
\end{subfigure}
\begin{subfigure}[t]{\linewidth}
    \center
    \includegraphics[width=\linewidth]{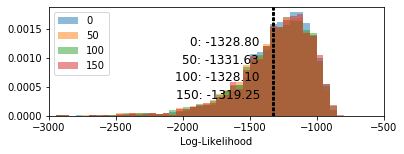}
    \caption{Glow, Gaussian mixture, Fashion-MNIST (1, 7)}
\end{subfigure}
\begin{subfigure}[t]{\linewidth}
    \center
    \includegraphics[width=\linewidth]{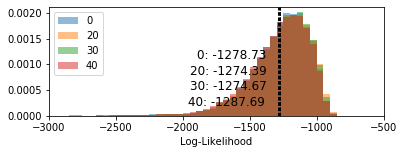}
    \caption{Glow, Generalized Gaussian mixture, Fashion-MNIST (1, 7)}
\end{subfigure}
\end{center}
\caption{Distances between two components
and the log-likelihoods assigned to
test data (Fashion-MNIST (1, 7)) by VAEs trained on Fashion-MNIST (1, 7).
The likelihoods assigned to the test data are relatively not affected
by the distance between two components.}
\label{fig:appendix-distance-fashion}
\end{figure}

Figure~\ref{fig:appendix-distance-mnist}, \ref{fig:appendix-distance-fashion}
show the histograms of the log-likelihoods assigned to
MNIST and the test data of Fashion-MNIST (label 1, 7) by 
Glow and VAEs trained on Fashion-MNIST (label 1, 7)
using different distances between two components.
Figure~\ref{fig:appendix-distance-mnist} shows the histograms
corresponding to the results reported in Figure~\ref{fig:distance-graph}.
The likelihoods assigned to the test data of Fashion-MNIST are not
affected by the distances between two components significantly
compared to those assigned to MNIST.

\subsection{unimodal Prior Models Trained on Simpler Data \label{appendix:unimodal}}
\begin{figure}[t]
\begin{center}
\begin{subfigure}[t]{\linewidth}
    \center
    \includegraphics[width=\linewidth]{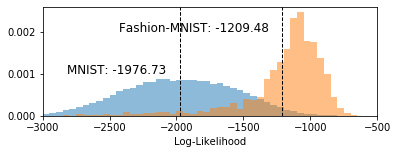}
    \caption{VAE, label 1}
\end{subfigure}
\begin{subfigure}[t]{\linewidth}
    \center
    \includegraphics[width=\linewidth]{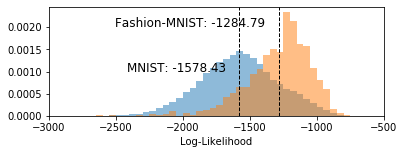}
    \caption{VAE, label 7}
\end{subfigure}
\begin{subfigure}[t]{\linewidth}
    \center
    \includegraphics[width=\linewidth]{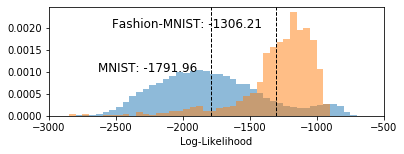}
    \caption{Glow, label 1}
\end{subfigure}
\begin{subfigure}[t]{\linewidth}
    \center
    \includegraphics[width=\linewidth]{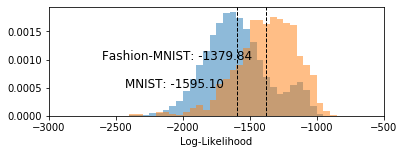}
    \caption{Glow, label 7}
\end{subfigure}
\end{center}
\caption{Histograms of the log-likelihoods assigned by models with standard Gaussian priors
trained on Fashion-MNIST (label 1 or 7).
The results corresponds to those in Table~\ref{tab:unimodal}.
While the models trained on a simpler data set assign lower likelihoods to
out-of-distribution inputs,
the models using multimodal distributions assign much lower likelihoods.}
\label{fig:unimodal}
\end{figure}

Figure~\ref{fig:unimodal} shows the log-likelihoods assigned to MNIST and the test data
by models with standard Gaussian priors trained on Fashion-MNIST (label 1 or 7).
Although the models assign lower likelihood to MNIST,
the effect of alleviating the out-of-distribution behaviour is not significant compared
to that of the model using a multimodal prior.

\subsection{Histograms of the Latent Variables \label{appendix:latentvariables}}
\begin{figure}[tb]
\begin{center}
\begin{subfigure}[t]{\linewidth}
    \center
    \includegraphics[width=\linewidth]{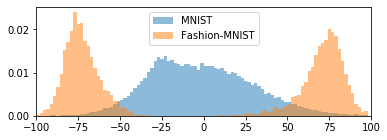}
    \caption{VAE}
\end{subfigure}
\begin{subfigure}[t]{\linewidth}
    \center
    \includegraphics[width=\linewidth]{figs/latentvariables/glow_twomodes_gaussian.png}
    \caption{Glow}
\end{subfigure}
\end{center}
\caption{Latent variables on the models
with bimodal Gaussian priors trained on Fashion-MNIST (label 1, 7).
The latent variables of MNIST reside in out-of-distribution areas.}
\label{fig:apendix-lv}

\begin{center}
\begin{subfigure}[t]{\linewidth}
    \center
    \includegraphics[width=\linewidth]{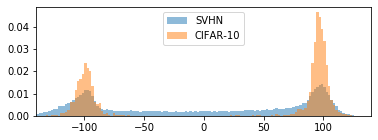}
    \caption{VAE}
\end{subfigure}
\begin{subfigure}[t]{\linewidth}
    \center
    \includegraphics[width=\linewidth]{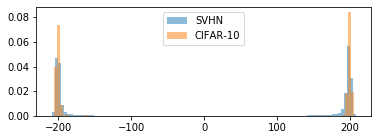}
    \caption{Glow}
\end{subfigure}
\end{center}
\caption{Latent variables on the models
with bimodal Gaussian priors trained on CIFAR-10 (label 0, 4).
The latent variables of SVHN reside near in-distribution areas.}
\label{fig:apendix-lv-cifar}
\end{figure}

\begin{figure}[tb]
\begin{center}
\begin{subfigure}[t]{\linewidth}
    \center
    \includegraphics[width=\linewidth]{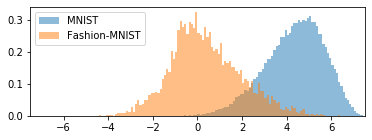}
    \caption{VAE, label 1}
\end{subfigure}
\begin{subfigure}[t]{\linewidth}
    \center
    \includegraphics[width=\linewidth]{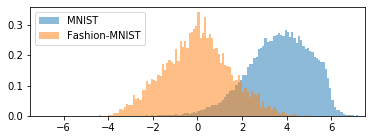}
    \caption{VAE, label 7}
\end{subfigure}
\begin{subfigure}[t]{\linewidth}
    \center
    \includegraphics[width=\linewidth]{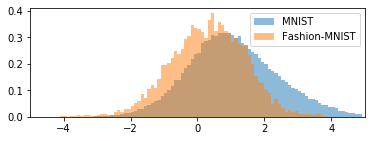}
    \caption{Glow, label 1}
\end{subfigure}
\begin{subfigure}[t]{\linewidth}
    \center
    \includegraphics[width=\linewidth]{figs/latentvariables/glow_label7_gaussian.png}
    \caption{Glow, label 7}
\end{subfigure}
\end{center}
\caption{Histograms of the latent variables on the models
with standard Gaussian prior trained on Fashion-MNIST label 1 or 7.
We select the dimension whose mean
of the latent variables of MNIST is farthest from zero.}
\label{fig:unimodal-lv}
\end{figure}

Figure~\ref{fig:apendix-lv} shows the histograms of the latent variables
of the models with bimodal prior distributions trained on Fashion-MNIST (label 1, 7).
For the results on VAE,
we show the histograms of the means of posterior distributions.
Figure~\ref{fig:unimodal-lv} shows the histograms of the latent variables on the models
with standard Gaussian priors trained on Fashion-MNIST (label 1 or 7).
In the latent variable spaces of the models with multimodal prior distributions,
MNIST resides in out-of-distribution areas
and it does not have a large overlap with Fashion-MNIST.
By contrast,
while we select the dimension whose mean of the latent variables of MNIST
is farthest from zero (the mean of the prior distributions),
the latent variables of MNIST have a large overlap with those of Fashion-MNIST
on the models with unimodal prior distributions, especially on Glow.

Figure~\ref{fig:apendix-lv-cifar} shows the histograms of the latent variables
of the models with bimodal prior distributions trained on CIFAR-10 (label 0, 4).
In contrast to Fashion-MNIST vs.\ MNIST,
the latent variables of SVHN are located near the in-distribution areas.

\end{document}